\documentclass{article}
\usepackage{spconf,amsmath,epsfig}
\usepackage{color}
\usepackage{multirow}
\usepackage{amssymb}
\usepackage[colorlinks,linkcolor=red]{hyperref}
\newtheorem{thm}{Theorem}

\begin{document}\sloppy

\def\x{{\mathbf x}}
\def\L{{\cal L}}

\title{Towards Digital Retina in Smart Cities: A Model Generation, Utilization and Communication Paradigm}
%

\name{Yihang Lou$^{1,4}$, Ling-Yu Duan$^{1,4}$, Yong Luo $^{1,4}$, Ziqian Chen$^{1,4}$, Tongliang Liu$^{2}$, Shiqi Wang$^{3}$, Wen Gao$^{1,4}$\thanks{Ling-Yu Duan is the corresponding author.}}
\address{National Engineering Lab for Video Technology, Peking University, Beijing, China$^1$\protect\\School of Computer Science, University of Sydney, Australia$^2$\protect\\School of Computer Science, City University of Hongkong, China$^3$\protect\\The Peng Cheng Laboratory, Shenzhen, China$^4$\protect\\$\{$yihanglou, lingyu, luoyong, wzziqian, wgao$\}$@pku.edu.cn, \protect\\tongliang.liu@sydney.edu.au, shiqwang@cityu.edu.hk}

\maketitle

\begin{abstract}
The digital retina in smart cities is to select what the City Eye tells the City Brain, and convert the acquired visual data from front-end visual sensors to features in an intelligent sensing manner. By deploying deep learning and/or handcrafted models in front-end devices, the compact features can be extracted and subsequently delivered to back-end cloud for search and advanced analytics.
In this context, we propose a model generation, utilization, and communication paradigm, aiming to address a set of unique challenges for better artificial intelligence services in smart cities. 
In particular, we present an integrated multiple deep learning models reuse and prediction strategy, which greatly increases the feasibility of the digital retina in processing and analyzing the large-scale visual data in smart cities.
The promise of the proposed paradigm is demonstrated through a set of experiments.
\end{abstract}
\section{Introduction}

Retina is the crucial and indispensable component of our visual system, which converts light signals into neuronal representations and acts as a filter that conveys specifically required and meaningful visual information to the brain \cite{wassle2004parallel}\cite{bao2018artificial}. 
In the retina, the rods and cones cells are responsible for the low and high light levels, as well as the color vision. The photosensitive retinal ganglion cells extract the complex features to complete the perception process. As such, retina not only perceives the visual information, but also works as a highly efficient visual data processing engine in the central nervous system. In analogous to the concept of retina, as illustrated in Fig. 1, the digital retina \cite{gao} that intelligently senses and processes the visual data in the City Brain is committed to revolutionize the artificial vision system of smart cities. 

In particular, the central processing unit in digital retina is feature extraction, which relies on deep learning and/or handcrafted models in the front-end visual sensors to directly extract and compress the features, such that the compact features can be efficiently sent to central server for visual analysis. To facilitate this process, the model, which is usually learned at the central server by leveraging the visual data, is the core component in the City Brain, 
and the learned models are subsequently delivered to the front-end devices for feature extraction and compression.
As such, the model generation, utilization and communication are essential in establishing the digital retina, especially in the sense that the collected visual data are featured by high variations in terms of locations, time and ambient environments. 
However, how the effective models can be feasibly generated by leveraging the existing massive models in different domains has not been fully exploited. 

 \begin{figure}[tbp]
\centering
  \includegraphics[width=0.95\linewidth]{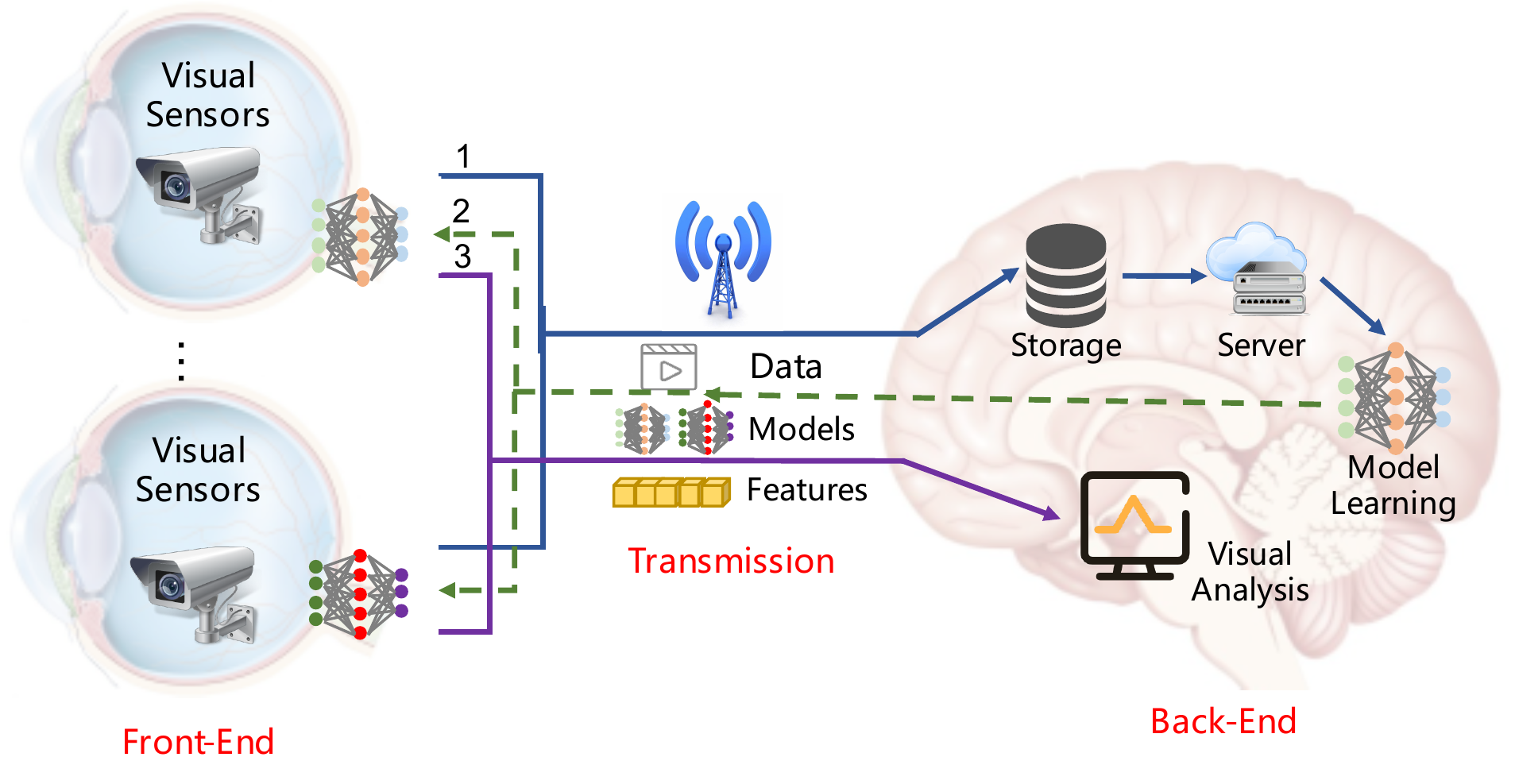}
  \vspace*{-10pt}
  \caption{Illustration of digital retina framework in smart cities. The data flow 1,2,3 denotes the visual data, models and feature transmission routines in digital retina, respectively.
  }
    \label{incremental_illustre}
    \vspace*{-15pt}
\end{figure}

In this paper, we focus on an integrated solution for effective model generation and efficient transmission by exploiting the cross-domain and inter-model relationships for the construction of digital retina. 
The transfer of learning has been studied for decades from the psychological \cite{woodworth1901influence} , and educational \cite{national2000people} perspectives, and it has been widely recognized that learning from prior experience or knowledge transfer can help human learning and improve performance. This motivates us to design a novel model transfer/reuse module that is particularly suitable in the proposed digital retina system. Besdies, the energy cost of the information transmission between neurons in human brain is usually very low due to the selective and adaptive regulation mechanism \cite{sterling2015principles}, this inspires us to propose a low transmission cost strategy that can deliver learning models incrementally and adaptively. 

Furthermore, with the constantly generated data in smart cities, the models undergo generation, updating, and distribution.
In particular, for the same tasks or similar tasks, there exists high correlations among different models even with different domains of training data. 
In view of this, we aim to efficiently make use of these models that operate the artificial intelligence applications in city brain from two perspectives. 
1) The existing models can be reused for model generation, even when the source and target models are in different domains. 2) These models can be efficiently delivered for better utilization and management.


In this work, we aim to explore more efficient and effective model utilization, management and distribution methodologies in the context of digital retina. 
The main contributions of this paper are summarized as follows:
\begin{itemize}
\item We propose a novel model generation, utilization and communication paradigm towards digital retina to better construct the artificial vision system in smart cities.
\item We explore an efficient model reuse strategy to learn more discriminative and domain adaptive models.
\item We develop a scalable model communication scheme by removing the inter-model redundancy to deliver the newly generated models at low transmission cost.
\end{itemize}

\section{Related Work}

\begin{figure*}[htbp]
\centering
  \includegraphics[width=0.92\linewidth]{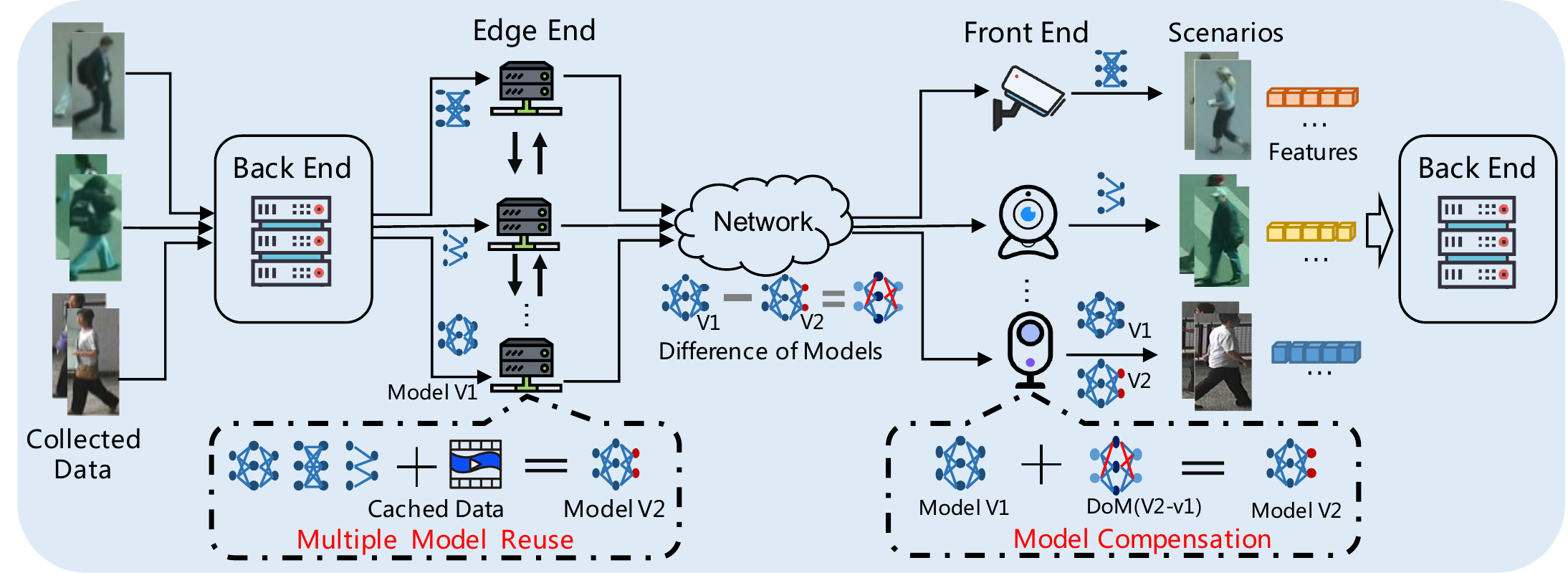}
 \vspace*{-5pt}
  \caption{Illustration of the proposed model generation, utilization and communication paradigm in digital retina.}
    \label{fig:framework}
    \vspace*{-10pt}
\end{figure*}

\textbf{Model Reuse.} Model reuse aims to explore the reusability of existing models for facilitating the model training in the target domain with limited resources and/or labeled data. Recently, a fixed model reuse (FMR) method was proposed in \cite{yang2017deep}, which assumes that a set of additional feature is provided for each training sample. Therefore, the method is actually feature reuse and more information contained in the deep models are ignored. Besides, only one type of source feature can be handled in \cite{yang2017deep}. A modal consistency multi-model reuse strategy was presented in \cite{xiangmodal}, but it assumes there are multiple modalities (e.g., features) in each domain, and cannot be used when there is only one modality. In addition, it only utilizes the labels of different models. Consequently, the abundant information contained in the models are ignored, and different source models should be presented in the testing phase. A bag of experts approach was proposed in \cite{jha2018bag} to directly reuse the output features or labels of some expert source models and hence suffers from the same drawbacks as in \cite{yang2017deep} or \cite{xiangmodal}. In this work, we propose a novel and general framework that can appropriately tackle these issues and make full use of the knowledge contained in multiple existing source models for training the target model, which greatly facilitates the model generation based on the models in the City Brain.

\textbf{Deep Neural Network Communication.} 
The deep neural network transmission aims to utilize and deliver the knowledge concentrated in the network model to facilitate different intelligent applications. 
In \cite{chen2018data}, the model compression is formulated from the perspective of transmission. As such, the redundancy among different models can be further exploited to facilitate many applications in front-end visual sensors. It is also shown that such scheme can be elegantly combined with the existing compression methods to form an integrated compression and communication framework.

\vspace{-13pt}
 
\section{Model Generation, Utilization and Communication Paradigm}
In this section, we demonstrate the model generation, utilization and transmission in the digital retina system of smart cities. 
As illustrated in Fig.\ref{fig:framework} , we propose an integrated solution with edge computing, which serves as an intermediate layer between the visual sensor and central cloud. The edge computing is effective in offloading the computation load from the central server and caching the data from the front-end visual sensors.
Accordingly, there arise multiple requirements for model generation and communication from different perspectives, which can be summarized as follows:


1) Model reuse between edge ends.
The edge nodes can utilize and reuse the models from the other edge nodes to perform particular tasks. 
However, it is widely acknowledged that there is severe domain bias of captured data in widely deployed visual sensors.
In real-world applications, the deployed front-end visual sensors in different locations may deal with the data in different domains. Regarding a specific task, multiple models trained in a particular domain can be utilized to generate a more discriminative model. 
For example, the trained models at several edge nodes can be transmitted to a particular edge node for multi-model reuse.

2) Model transmission from edge to front-end.
For the front-end, the deployed models from the edge nodes are often updated. 
In such scenario, there exist high correlations among a series of updating models. To economically deliver the newly generated models, the exploration of inter deep learning model redundancy removal is able to greatly reduce the transmission cost of the to-be-deployed models.


\section{Multiple Model Reuse and Prediction}

\subsection{Multi-Model Reuse}
In real-world applications, domain can be characterized by target characteristics, geospatial information and capture conditions. There usually exists domain bias for data distributions. Due to such domain gap, the model trained in one specific domain usually cannot well generalize to other domains. In the smart artificial vision systems, there often exist similar models for the same task,
which motivates us to leverage existing multiple models to obtain a domain specific one. Moreover, the acquired and cached data could also be leveraged, especially from a wide range of local visual sensors. Given these considerations, we propose a novel multi-model reuse strategy to improve the performance, especially for front-end visual sensors in a particular domain.

Given one target domain and $M$ source domains, we suppose there are a few labeled samples $\{x_n, y_n\}_{n=1}^{N_l}$ in the target domain. In the $m$-th source domain, a deep learning model $f_m(\Theta_m)$ is already trained using abundant labeled data.
The ultimate goal of multiple model reuse is to learn a model $f_T(\Theta_T)$ in the target domain using the pre-trained source models and limited labeled data in the target domain.
\begin{figure}[htbp]
\centering
  \includegraphics[width=0.8\linewidth]{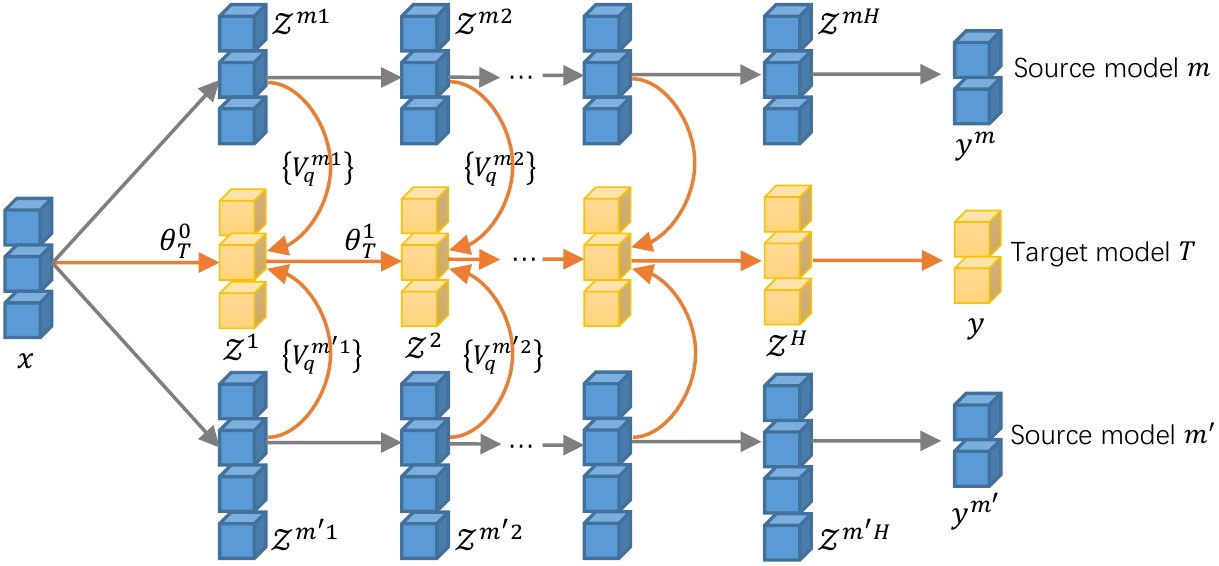}
  \caption{Architecture of the multi-model reuse framework.}
  \label{fig:model_architecture}
  \vspace*{-10pt}
\end{figure}

The architecture of the proposed multi-model reuse framework is shown in Fig.~\ref{fig:model_architecture}. In particular, we make a mild assumption that the pre-trained source models and target model are characterized by Convolutional Neural Networks (CNNs). To conduct reliable model reuse, we also assume that there are large amounts of unlabeled data $\{x_n\}_{n=N_l+1}^{N_l+N_u}$ in the target domain. Then for each (labeled or unlabeled) input $x$, we map it as the target hidden layer representation $\mathcal{Z}^h$ and multiple source hidden layer representations $\mathcal{Z}^{mh}, m=1,\ldots,M$, where $h$ is the layer index and $M$ is the number of source models. Here, each representation $\mathcal{Z}$ is an $Q$-order tensor of size $d_1 \times d_2 \times \ldots \times d_Q$. To reuse the multiple source models, different ${\mathcal{Z}^{mh}}$ are mapped into a common subspace, and $\mathcal{Z}^h$ is enforced to be close to the common representation. This is achieved by employing the multi-view learning strategy \cite{long2008general}, \textit{i.e.}, using $\mathcal{Z}^h$ to reconstruct each ${\mathcal{Z}^{mh}}$. As such, $\mathcal{Z}^h$ can be regarded as a meta-embedding of the different source layer representations. In this way, all the features in the source domains are utilized to improve the hidden layer representation, leading to a better model compared to the one trained using only the limited labeled information in the target domain.
The objective function is given by,
\begin{equation}
\begin{split}
\epsilon(\Theta_T; x_n, y_n) = & \frac{1}{N_l} \sum_{n=1}^{N_l} L(f_T(\Theta_T;x_n),y_n) \\
& + \gamma \sum_{n=1}^{N_l+N_u} \sum_{h=1}^{H'<H} R(\mathcal{Z}_n^h; \{\mathcal{Z}_n^{mh}\}),
\end{split}
\label{eq:objective}
\end{equation}
where $\Theta_T = \left \{ \theta_T^h \right \}^H_{h=0} $ is the set of all parameters of the target learning task, $\mathcal{Z}_n^h = \varphi (\theta_T^{h-1}; \mathcal{Z}_n^{h-1})$ and $\varphi(\cdot)$ is an activation function; $L(\cdot)$ is any loss that is adopted in deep learning and $R(\cdot)$ is a regularization term that enables model reuse. A general formulation of $R(\cdot)$ can be given by,
\begin{equation}
\begin{split}
& R(\mathcal{Z}_n^h; \{\mathcal{Z}_n^{mh}\}) \\
& = \sum_{m=1}^M \alpha_m \left \| \mathcal{Z}_n^{mh} - \mathcal{Z}_n^h \times_1 V_1^{mh} \ldots \times_Q V_Q^{mh} \right \|_F^2,
\end{split}
\label{eq:regularization_tensor}
\end{equation}
where each $V_q^{mh}$ is a transformation matrix of size $d_q^{mh} \times d_q^{h}$, $\times_q$ is the $q$-mode tensor-matrix product, and $\{\alpha_m\}$ are the weights that reflect the importance of different source models and satisfy $\sum_m \alpha_m = 1$. When the hidden layer representation is a vector, \textit{i.e.}, $\mathcal{Z}$ is an one-order tensor ($Q=1$), the regularization term becomes,
\begin{equation}
\begin{split}
R(\mathbf{z}_n^h; \{\mathbf{z}_n^{mh}\}) = \sum_{m=1}^M \alpha_m \left \| \mathbf{z}_n^{mh} - V^{mh} \mathbf{z}_n^h \right \|^2_2.
\end{split}
\label{eq:regularization_vector}
\end{equation}
The additional parameters $\{V^{mh}\}$ would increase the risk of over-fitting. In practice, this is not an issue since $\{V^{mh}\}$ can be learned using the large amounts of unlabeled data.

\subsection{Theoretical Analysis}
For simplicity, we treat the deep learning models as end-to-end learning algorithms. Specifically, we denote the prediction function for the target model as $f_T$, while the prediction functions for the source models as $f_m, m=1,\ldots,M$. To measure the accuracy of a model $f$, we introduce the expected risk
$\mathcal{R}(f)=\mathbb{E}_{(x,y)\sim D}[L(f(x),y)]$. Then we have the following theorem, which shows that the proposed model can benefit from the source models with a theoretical guarantee on the expected risk (The least square loss is simply adopted). 
\begin{thm}
Assume the linearity of their input-output map
, \textit{i.e.}, $(f_1\pm f_2)(x)=f_1(x)\pm f_2(x)$ and $f(x)\leq \|f\|\|x\|$, and the feature space is bounded, \textit{i.e.}, $\|x\|\leq r$, then when adopting the least square loss, we have 
\begin{equation}
\mathcal{R}(f_{T,N_l}) \leq 2M \sum_{m=1}^{M} \alpha_m^2 \mathcal{R}(V^m f_m)(r^2/\gamma + 1).
\end{equation}
\end{thm}
Here, $\mathcal{R}(f_{T,N_l})$ is the expected risk of the target model trained using the $N_l$ labeled samples, and $\mathcal{R}(V^m f_m)$ is the risk of the transformed source model w.r.t. the data in the target domain. We leave the proof in the supplementary material due to the limited space.\footnote{The supplementary material is available at \url{https://github.com/PKU-IMRE/Retina}.} Since the source models are well-trained (such as using abundant labeled data), we believe that there exists (or we can learn) some $V^m$ such that $\mathcal{R}(V^m f_m)$ is small if the source tasks are related to the target task. When $\{\alpha_m\}$ and $\{V^m\}$ are determined appropriately, the term $M \sum_{m=1}^{M} \alpha_m^2 \mathcal{R}(V^m f_m)$ can be small. So the expected risk $\mathcal{R}(f_{T,N_l})$ of the target model is guaranteed to be low. We do not have such guarantee when the designed regularization term does not exist, since the expected risk of the target model would be high, in which the model is prone to over-fitting for limited labeled data in the target domain. 

\subsection{Model Prediction}
Based on the proposed methodology, the generated models deployed on front-end visual sensors will be often updated. 
In this context, the model communication acts as an essential component in the City Brain. In particular, with the philosophy of model reusing, it is painlessly to obtain the models frequently, such that efficient communication of these models is highly desired.
Hence, 
we investigate the economic model communication based on the difference of models (DoM) between the existing model (e.g., existing in both sender and receiver) and the to-be-transmitted model.
We compute the DoM between the prediction and the to-be-compressed models by computing the difference for each corresponding weight layer-by-layer. Denote the prediction model as $M_p$ and the to-be compressed model as $M_c$, then the DoM between $M_p$ and $M_c$ can be computed as follows,
\begin{equation}
\begin{aligned}
M_{DoM}(h,i) = M_c{(h,i)} - M_p{(h,i)},
\end{aligned}
\label{margin}
\end{equation}
where $h$ and $i$ signify the layer index and weight index in each layer, respectively. Subsequently, the weight differences \(w\) 
are quantized to \(q_w\) according to scalar quantization,
\begin{equation}
q_{w_{DoM}}(h,i) =\text{r}\left(\frac{w_{DoM}(h,i) \times10^{s\_bits}+f\times10^{q\_bits}}{10^{q\_bits}}\right).
\end{equation}
Here, \(s\_bits\) is the amplification of the weights, and \(q\_bits\) determines the degree of quantization. The higher \(q\_bits\) indicates the less coding bits. $r()$ denotes the round operation. 
In the receiver end, the \(q_w\) is further de-quantized to recover the weight \(\hat{w}\) as follows,
\begin{equation}
\hat{w}_{DoM}(h,i) = (q_{w_{DoM}}(h,i)\times 10^{q\_bits-s\_bits} ).
\end{equation}
Then model compensation is performed to recover the model that is desired to be transmitted, \textit{i.e.},
\begin{equation}
\begin{aligned}
\hat{M}_{c}(h,i) = \hat{w}_{DoM}(h,i) + M_p{(h,i)}.
\end{aligned}
\label{margin}
\end{equation}
In addition, the recovered source models from model prediction are also allowed to facilitate the model training in the target domain through the multi-model reuse method.

\section{Experimental Results}
\subsection{Experimental setup}
We conduct experiments on a typical task of person ReID \cite{sun2017beyond} in smart city applications, which aims to find person images in the database as the query person image.
The reason for choosing this task is that the variant capture conditions and data collection scenarios result in severe domain bias in data distribution. Four different person ReID datasets Duke \cite{zheng2017unlabeled}, Market1501 \cite{zheng2015scalable}, MSMT17 \cite{wei2017person}, CUHK03 \cite{li2014deepreid} are used in experiments, and the details of them are shown in Table \ref{dataset_info}. In all experiments, the reused models are trained on MSMT, CUHK03 and Market1501 datasets, and the target model is trained and tested on Duke dataset.
\vspace{-10pt}
\begin{table}[h]
\centering
\caption{Details on the datasets in our experiments.}
\label{dataset_info}
{\small
\resizebox{0.78\columnwidth}{!}{
\begin{tabular}{|c|c|c|c|c|}
\hline
    Dataset  & images/IDs & train & test \\ \hline
    Duke \cite{zheng2017unlabeled}  &  36,411/1,812 & 16,522/702 & 19,919/1,110  \\ \hline
    Market1501 \cite{zheng2015scalable}   & 32,688/1,501 & 12,936/751  & 19,752/750 \\ \hline 
      MSMT17 \cite{wei2017person} &  126,441/4,101 & 32,621/1,041 & 93,820/3,060 \\ \hline
     CUHK03 \cite{li2014deepreid}  &  28,192/1,467 & 26,264/1,367 & 1,928/100 \\ \hline
\end{tabular}
}}
\vspace{-10pt}
\end{table}

\textbf{Network Architecture.} We adopt the ResNet50 network \cite{he2016deep} as our base network, and use softmax loss as supervision.

\textbf{Evaluation Metrics.} The person ReID task is regarded as a retrieval task. The mean Average Precision (mAP) and Top1 Accuracy are used for evaluation.

\textbf{Unlabeled Data Usage.} To perform the unlabeled data experiments, we split the above training set into two parts, \textit{\textit{i.e.}}, the labeled and unlabeled. In the unlabeled part, the label information are not used in the experiments. For example, 30\% Duke means only 30\% samples are used as labeled samples and the rest serve as unlabeled samples during training.

\vspace{-10pt}
\subsection{Results of Model Reuse}
\vspace{-5pt}
\begin{figure}[htbp]
\begin{tabular}{@{}c@{} @{}c@{}@{}c@{} @{}c@{}}
	\includegraphics[width=1.7in]{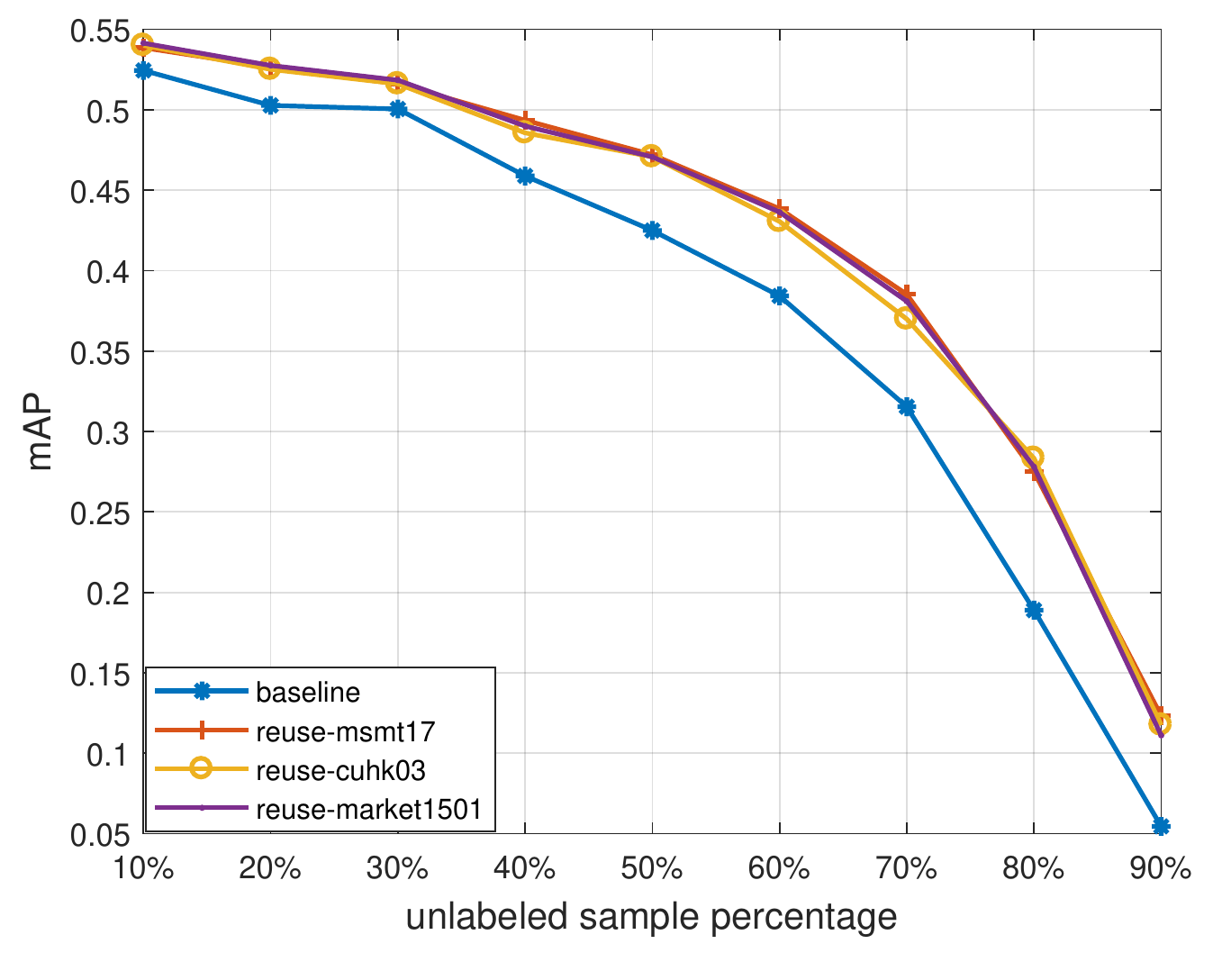} &
	\includegraphics[width=1.75in]{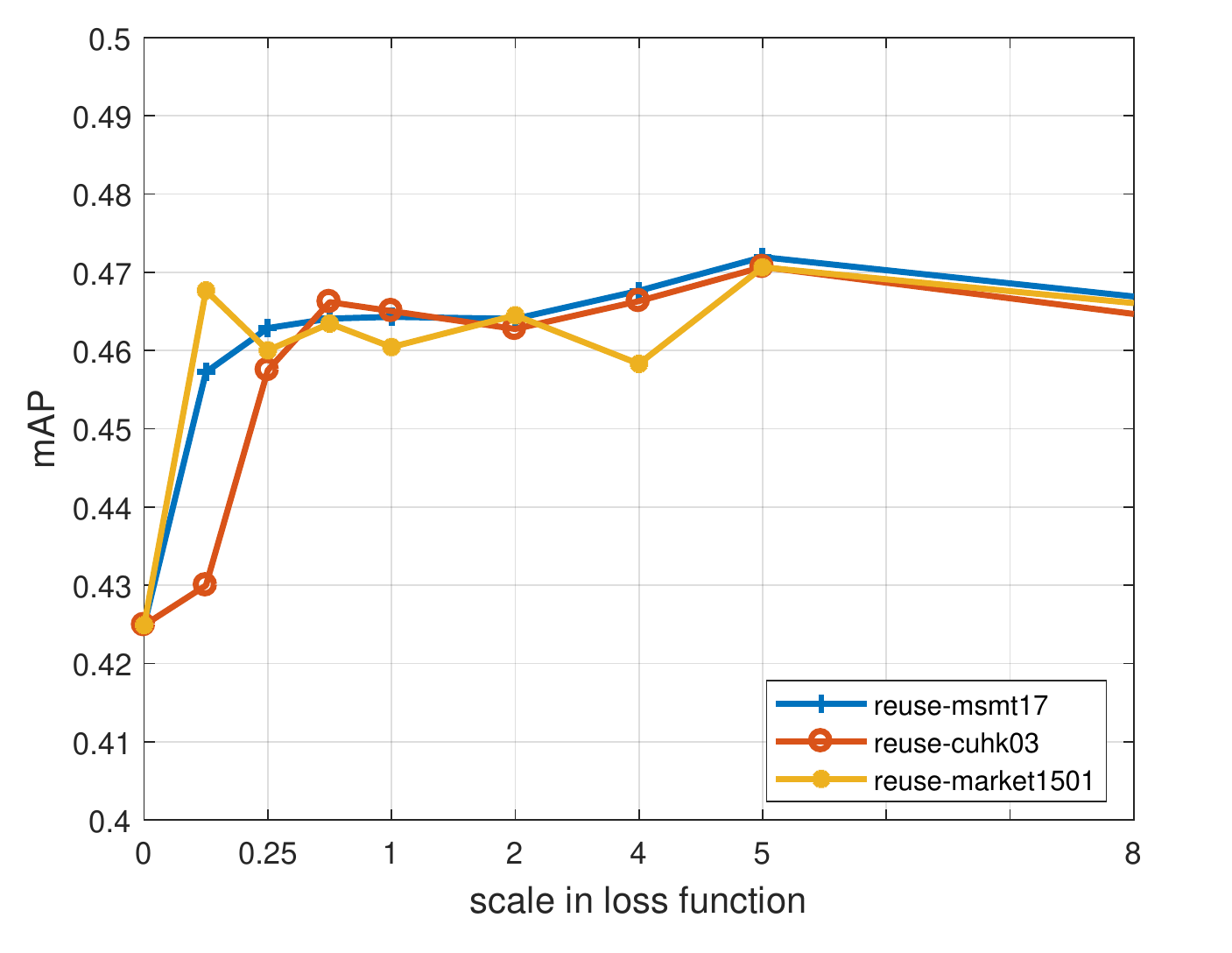}  & \\
	{(a)} & {(b)}  \\
\end{tabular}
\vspace{-10pt}
\caption{The model reuse performance on Duke test set. (a) The performance of reusing different single models on Duke train set by varying the percentage of unlabeled data; (b) The performance by setting different scale $\gamma$ in the loss function.} 
\label{unlabel_single1}
\vspace{-15pt}
\end{figure}
\vspace{-5pt}
\begin{table}[htbp]
\centering
\caption{The mAP performance by setting different scale $\gamma$ when reusing two models on the Duke test set. (70\% samples in Duke train set are used as unlabeled samples)}
\label{two_model_map}
\centering
{\resizebox{0.78\columnwidth}{!}{
\begin{tabular}{|c|c|c|c|}
\hline
    Scale & MSMT+CUHK &Market+CUHK & MSMT+Market\\
    \hline
    $\gamma$=1 & 37.91 & 39.19& 39.63 \\
    $\gamma$=2& 37.42  & 39.23 & \textbf{40.16} \\
    $\gamma$=4 & 39.37&  \textbf{39.43} &38.70				\\
    $\gamma$=5 & \textbf{39.39} & 39.22 & 38.96\\
    $\gamma$=8 & 39.33& 38.49 & 39.18 \\
    $\gamma$=16 & 38.37&38.20 &38.84  \\
    \hline
    baseline & MSMT only & CUHK only & Market only \\
    31.53 & 38.52& 36.98 & 38.09 \\ 
\hline
\end{tabular}}
}
\vspace{-10pt}
\end{table}

\textbf{Single Model Reuse.} We first demonstrate the performance of single model reuse in Fig.~\ref{unlabel_single1}(a)$\&$(b). The baseline model is trained with the labeled training part of Duke dataset. Three different models trained on MSMT17, CUHK03 and Market1501 are used for model reuse and the target model is trained and tested on the Duke dataset. It can be observed that additional model reuse significantly boosts the performance of ReID model over the baseline. Moreover, with different settings on the percentage of unlabeled data, the model trained by reusing strategy can consistently outperform the baseline. 

\textbf{The hyper-parameter analysis.}
The $\gamma$ balances the empirical loss and regularization term for reuse in the training objective. Properly choosing the value of $\gamma$ can improve the performance. To investigate the sensitivity of model w.r.t. $\gamma$, we vary the $\gamma$ and the performances are shown in Fig.~\ref{unlabel_single1}.
Our model remains fairly stable across a wide range of $\gamma$ from $2^{-3}$ to $2^3$. Besides, in Table \ref{two_model_map}, we present the performances of reusing two models on different scales. It can be observed that the overall performances under different scales are close. As such, we select $\gamma=5$ in the following experiments. 


\textbf{Multi-Model Reuse.} The results of multi-model reuse are shown in Table~\ref{multi_model}. Compared to the baseline models, reusing additional models achieves better performance both in mAP and Rank1. Moreover, the incremental performance gains can be consistently achieved by increasing the number of reused models. With three reused models, we can achieve 41.2\% mAP, which significantly outperforms the baseline 31.5\% mAP. We also compare with the state-of-the-art methods such as PCB \cite{sun2017beyond}, which has superior performance over our adopted softmax baseline. 
With incremental multi-model reuse, our baseline model has been significantly improved. 
When incorporating three models into the model reusing, we outperform the PCB \cite{sun2017beyond} by about 5\% mAP and 4\% Rank 1.\footnote{The models trained with multi-model reuse strategy are available at \url{https://github.com/PKU-IMRE/Retina}.} 
\vspace*{-10pt}
\begin{table}[htbp] 
\centering
\caption{The performance comparison and the incremental gain by increasing reused models on the Duke test set. (70\% samples in Duke train set are used as unlabeled samples)} 
\label{multi_model}
\centering
{\resizebox{0.65\columnwidth}{!}{ 
\begin{tabular}{|c|c|c|}
\hline
    Model  & mAP & Rank-1\\
    \hline
    Triplet \cite{wei2017person}&34.31 &54.30 \\
    PCB \cite{sun2017beyond} & 36.62 & 57.05 \\
    DefenseTriplet \cite{hermans2017defense} & 35.96 & 55.97 \\
    AlignedReID\cite{zhang2017alignedreid} & 35.35 & 55.38 \\
    AlignedReID+Mutual Learning\cite{zhang2017alignedreid} &36.60 &55.48 \\  \hline
    Softmax Baseline & 31.53 & 49.55 \\
    +MSMT & 38.52 & 58.61  \\
    +CUHK & 36.98 &  58.34			\\
    +Market & 38.09 & 57.40 \\
    +MSMT+CUHK & 39.93 & 59.87  \\
    +Market+CUHK & 39.22 & 60.18 \\
    +MSMT+Market & 40.16& 59.87 \\
    \hline
    +MSMT+CUHK+Market & 41.24 & 61.04 \\
\hline
\end{tabular}}
}
\vspace*{-15pt}
\end{table}

\subsection{Model Prediction}
\vspace*{-15pt}
\begin{table}[htbp]
\caption{The mAP performance comparison over Duke test set by transmitting models with/without DoM subject to different compression bits (\textit{i.e.}, different quantization levels).}
\newcommand{\tabincell}[2]{\begin{tabular}{@{}#1@{}}#2\end{tabular}}
\footnotesize
\centering
\resizebox{0.88\columnwidth}{!}{
\begin{tabular}{|c|c|c|c|}
\hline
\multirow{2}{*}{Model} & DoM(V1-V0)  & DoM(V2-V1)  & DoM(V3-V2)  \\ \cline{2-4} 
                                  & Model-V1     & Model-V2     & Model-V3     \\ \hline
\multirow{2}{*}{original}          & 39.42 & 40.27 & 41.24 \\ \cline{2-4} 
                                  & 39.42 & 40.27& 41.24\\ \hline
\multirow{2}{*}{compression bits=7}         & 39.42 & 40.27  & 41.24 \\ \cline{2-4} 
                                  & 39.42 & 40.27 & 41.24 \\ \hline
\multirow{2}{*}{compression bits=6}         & 39.42 & 40.27 & 41.25 \\ \cline{2-4} 
                                  & 39.41 & 40.26 & 41.23 \\ \hline
\multirow{2}{*}{compression bits=5}         & 39.44 & 40.25 & 41.25 \\ \cline{2-4} 
                                  & 39.36 & 40.19 & 41.17 \\ \hline 
\multirow{2}{*}{compression bits=4}         & 39.36 & 39.76 & 40.73 \\ \cline{2-4} 
                                  & 38.54& 39.02 & 40.06 \\ \hline
\multirow{2}{*}{compression bits=3}         & 33.64 & 36.48   & 39.85 \\ \cline{2-4}                                     & 0.16   & 0.16 & 0.16   \\ \hline                                  
\end{tabular}}
\label{dom}
\vspace*{-10pt}
\end{table}

The enhanced model produced by multi-model reuse is deployed to the front-end by the incremental model updating strategy.
We present the performance in Table \ref{dom} with the decrease of compression bits ($s\_bits - q\_bits$). The smaller compression bits means coarser quantization, and the \(s\_bits\) is set to 12 and $f$ is set to 0.3. 
Moreover, the model v1/v2/v3 represent the different versions of the same models with incrementally better performance.
Given the same to-be-compressed models, higher compression ratios and better performances can be achieved with the DoM, say compression bits = 5 to 3. For better understanding of Table \ref{dom}, we plot the compression ratio changes in terms of different quantization with/without DoM in Fig.~\ref{incremental_illustre}. The DoM strategy significantly outperforms the simple single model compression scheme. 

The model prediction is able to consistently improve the performance by using model sharing information. When compression bits = 3, the DoM strategy can well maintain the performance (39.42$\rightarrow$33.64) while the single model compression strategy even collapses (39.42$\rightarrow$0.16). The reason is that DoM compresses the differences of the models while the single model compression is applied to the whole model. Thus, the DoM is more suitable for delivering incremental information under constrained transmission environment.

 \vspace*{-10pt}
\begin{figure}[htbp]
\centering
  \includegraphics[width=0.63\linewidth]{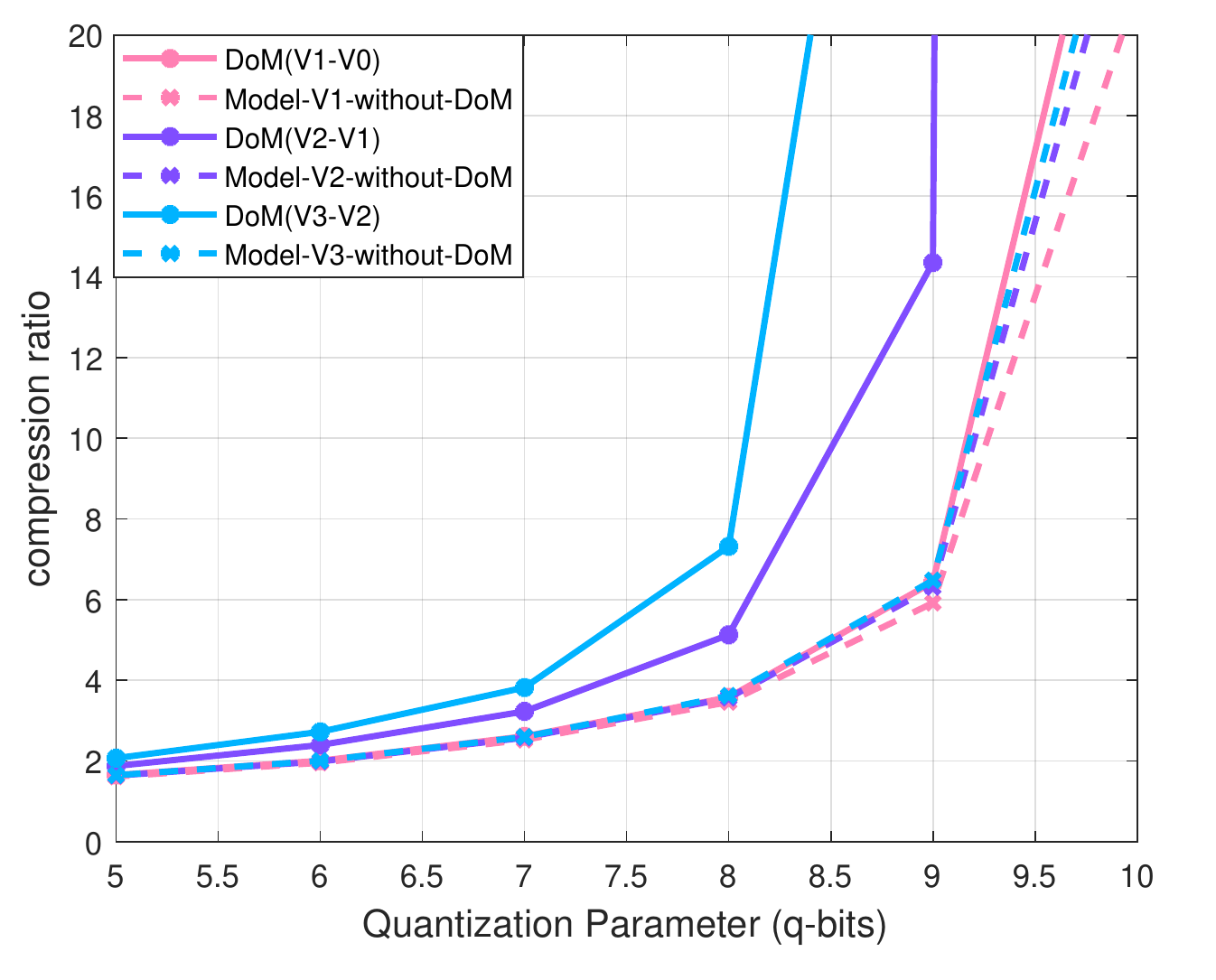}
  \vspace*{-15pt}
  \caption{Comparisons of the compression ratios between with/without DoM under different quantization levels. }
  \vspace*{-20pt}
    \label{incremental_illustre}
\end{figure}
\vspace{-5pt}
\section{Conclusion}
In this work, we show the potentials of model generation, utilization and communication in constructing the digital retina for artificial intelligence applications in smart cities. To demonstrate the benefits of adopting model reuse and prediction, we use the challenging problem of person ReID to reveal that properly reusing models is effective to deal with the data collected in a wide range of visual sensors. In the future, we will systematically integrate model generation, utilization, communication and standardization for establishing the intelligent, economic and efficient digital retina in smart cities.
\vspace{-14pt}
\paragraph*{Acknowledgement:}
This work was supported by the National Natural Science Foundation of China under Grant 61661146005 and Grant U1611461, and in part by the National Basic Research Program of China under Grant 2015CB351806, and in part by Australian Research Council Project DE-1901014738, and in part by Hong Kong RGC Early Career Scheme under Grant 9048122 (CityU 21211018).


\bibliographystyle{IEEEbib}
\bibliography{icme2019template}

\end{document}